\documentclass[letterpaper, 10 pt, conference]{ieeeconf}  % Comment this line out if you need a4paper

\IEEEoverridecommandlockouts                              % This command is only needed if 
\usepackage{caption}
\usepackage{graphicx}
\usepackage{stfloats}
\usepackage{float} 
\usepackage{subcaption}
\usepackage{color}
\usepackage{amsmath}
\usepackage{cite}
\usepackage{makecell}
\usepackage{booktabs}
\usepackage{multirow}
\usepackage{diagbox}
\makeatletter
\let\NAT@parse\undefined
\makeatother
\usepackage{hyperref}
\hypersetup{pdfstartview=FitH,
            colorlinks=true,
            linkcolor=red,
            anchorcolor=blue,
            citecolor=green
            }
\usepackage{colortbl}
\usepackage{amsfonts}
\usepackage[table]{xcolor}
\usepackage{soul}
\newcommand{\ignore}[1]{}

\newcommand{\myAbl}{\cellcolor{gray!20}}
\definecolor{cGreen}{RGB}{100,180,100}
\definecolor{cRed}{RGB}{220,50,0}
\definecolor{Klein_Blue}{rgb}{0.0, 0.129, 0.6}

\newcounter{RNum}
\renewcommand{\theRNum}{\arabic{RNum}}
\newcommand{\Remark}{\noindent\textit{\textbf{Remark}~\refstepcounter{RNum}\textbf{\theRNum}: }}

\soulregister\Remark7

\begin{document}
\title{\LARGE \bf
CGTrack: Cascade Gating Network with Hierarchical Feature Aggregation for UAV Tracking}
\author{
Weihong Li$^{1}$, Xiaoqiong Liu$^{2}$, Heng Fan$^{2,\dagger}$, and Libo Zhang$^{3,\dagger,*}$ % <-this % stops a space
\thanks{$^{\dagger}$Equal advising and co-last authors; $^*$Corresponding author}
\thanks{$^{1}$Weihong Li is with the Hangzhou Institute for Advanced Study, University of Chinese Academy of Sciences, Hangzhou 310024, China.}%
\thanks{$^{2}$Xiaoqiong Liu and Heng Fan are with the Dept. of Computer Science and Engineering, University of North Texas, Denton, TX 76207, USA.}%
\thanks{$^{3}$Libo Zhang is with the Institute of Software Chinese Academy of Science, Beijing 100190, China. {\tt\small libo@iscas.ac.cn}}%
}

% \captionsetup{font={small}}

\maketitle
\thispagestyle{empty}
\pagestyle{empty}

%%%%%%%%%%%%%%%%%%%%%%%%%%%%%%%%%%%%%%%%%%%%%%%%%%%%%%%%%%%%%%%%%%%%%%%%%%%%%%%%
\begin{abstract}
    Recent advancements in visual object tracking have markedly improved the capabilities of unmanned aerial vehicle (UAV) tracking, which is a critical component in real-world robotics applications. While the integration of hierarchical lightweight networks has become a prevalent strategy for enhancing efficiency in UAV tracking, it often results in a significant drop in network capacity, which further exacerbates challenges in UAV scenarios, such as frequent occlusions and extreme changes in viewing angles. To address these issues, we introduce a novel family of UAV trackers, termed CGTrack, which combines explicit and implicit techniques to expand network capacity within a coarse-to-fine framework. Specifically, we first introduce a Hierarchical Feature Cascade (HFC) module that leverages the spirit of feature reuse to increase network capacity by integrating the deep semantic cues with the rich spatial information, incurring minimal computational costs while enhancing feature representation. Based on this, we design a novel Lightweight Gated Center Head (LGCH) that utilizes gating mechanisms to decouple target-oriented coordinates from previously expanded features, which contain dense local discriminative information. Extensive experiments on three challenging UAV tracking benchmarks demonstrate that CGTrack achieves state-of-the-art performance while running fast. Code will be available at \href{https://github.com/Nightwatch-Fox11/CGTrack}{https://github.com/Nightwatch-Fox11/CGTrack}.
\end{abstract}

%%%%%%%%%%%%%%%%%%%%%%%%%%%%%%%%%%%%%%%%%%%%%%%%%%%%%%%%%%%%%%%%%%%%%%%%%%%%%%%%
\section{Introduction}
Unmanned aerial vehicles (UAVs) commonly refer to drones remotely operated by a human operator without any pilot on board. The rapid development of UAVs has boosted numerous real-world applications, \textit{e.g.}, logistic and product deliveries~\cite{UAV_logistic}, UAV-assisted IoT applications~\cite{UAV_iot}, and robotic automation~\cite{UAV_auto}. Despite the remarkable progress in visual object tracking, achieving efficient and accurate UAV tracking remains fraught with significant challenges, such as frequent scale changes, extreme viewing angles, and severe occlusions. These issues are particularly pronounced in the context of fast-moving drones. Therefore, it is crucial to develop more robust and efficient network designs, especially for edge devices with limited power resources.

In general, the majority of UAV trackers can be categorized into two types: discriminative correlation filters (DCF)-based trackers~\cite{kcf,danelljan2017eco,ATOM,DiMP,PrDiMP,KYS} or deep learning (DL)-based trackers~\cite{SiameseFC,SiameseRPN,SiamBAN,Stark,TransT,ostrack,SeqTrack}. Despite the superior efficiency brought by Fourier transformation, the accuracy of DCF-based trackers has fallen far behind that of the DL-based trackers. Among modern DL-based trackers, those based on Siamese networks~\cite{SiameseFC,SiameseRPN,SiamBAN,SiamFC++,SiamCAR,SiamRPNplusplus,SiamMask} are the most prevalent. They employ the strategy of ``divide-and-conquer'', where the template and the search region features are extracted separately before relation modeling. However, as described in ~\cite{ostrack}, the features extracted by Siamese networks lack essential target-oriented discriminative information, resulting in significant performance degradation in UAV scenarios, particularly during high-speed movement. Recently, Transformer~\cite{2017Attention} has played a pivotal role in the field of UAV tracking~\cite{hift,SGDViT,ClimRT} due to its superior capacity of modeling global relationships. Moreover, with the power of pre-trained Vision Transformer (ViT) models~\cite{ViT,clip21,CAE,dinov1,dinov2}, one-stream frameworks~\cite{ostrack,simtrack,rom,HiT} exhibit superior performance in both accuracy and efficiency when compared to Siamese-based trackers. However, Transformer-based trackers are burdened by high computational costs brought by ViTs and often neglect critical local information\cite{Conformer}, leading to failures in extreme UAV scenarios.

\begin{figure}[tbp]%调节图片位置，h：浮动；t：顶部；b:底部；p：当前位置 
 \centering
 \setlength{\abovecaptionskip}{1pt}
 \includegraphics[width=1\linewidth]{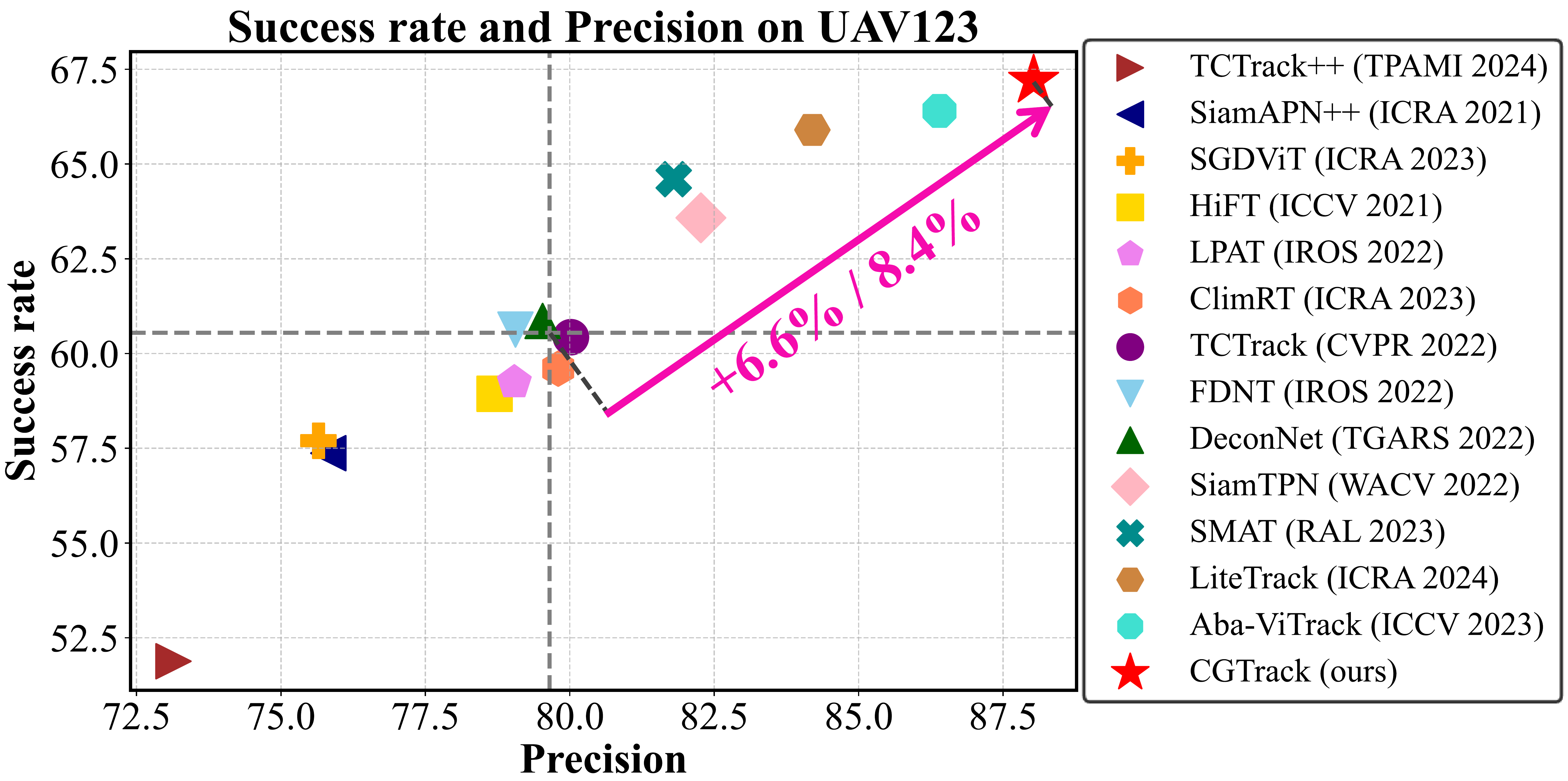}
 \caption
 {
    Comparison of success rate and precision between CGTrack and other 13 state-of-the-art (SOTA) trackers on the authoritative UAV123 benchmark~\cite{uav123}. CGTrack achieves SOTA performance in both precision and success rate, surpassing the average performance of 13 trackers by \textbf{6.6\%} and \textbf{8.4\%} respectively. Best viewed in color for all figures in this paper.
 }
 \label{fig:title}
 \vspace{-7mm}
\end{figure}

In this work, we address the aforementioned challenges by introducing a lightweight, hierarchical one-stream tracking framework. By adopting a lightweight hierarchical ViT as backbone, we obtain hierarchical features that preserve rich global contextual information. To effectively enhance network capacity with these hierarchical features, we propose a Hierarchical Feature Cascade (HFC) module, inspired by DenseNet~\cite{densenet}, which highlights the strength of feature reuse. However, unlike the original DenseNet~\cite{densenet}, our HFC module simplifies dense connections into a cascade structure by scaling multi-level features to a uniform size and later concatenating them. This approach allows us to obtain a feature map containing both deep semantic information and shallow detail information without additional parameters or FLOPs. The HFC module explicitly increases the network width (\textit{i.e.}, channel number) to provide rich contextual information for subsequent fine-grained discriminative feature extraction. The enriched feature map contains both discriminative local details and global context which are particularly crucial in resource-constrained challenging UAV scenarios. Additionally, we introduce a Residual Squeeze-and-Excitation (SE) module to apply coarse-grained gating on the feature maps after each concatenation, improving gradient flow through its residual design. 

To fully leverage the feature map generated by the HFC module, we improve the center head design in modern trackers~\cite{ostrack,Gao2023GeneralizedRM, Zhu2023VisualPM, Cai2023HIPTrackVT} by proposing the Lightweight Gated Center Head. Inspired by~\cite{StarNet}, we replace the basic Conv-BN-ReLU (CBR) block with an Efficient Gating (EG) block. The EG block first maps the features into a high-dimensional nonlinear feature space, where gating is performed subsequently via the Hadamard product. The gating mechanisms have shown superior capability in enhancing local fine-grained details in the field of image inpainting~\cite{deepfillv2}. In light of this, we employ EG block to further mine local discriminative information which is critical for addressing the challenges inherent in UAV scenarios.

In summary, our contributions in this paper are as follows:
\begin{itemize}
	\item  We propose CGTrack, a family of UAV tracking architecture aiming at combining global context provided by lightweight ViT with mined local discriminative details from hierarchical features to achieve robust UAV tracking.
	
	\item  Leveraging the art of feature reuse, a novel HFC module is presented whereby hierarchical features are aggregated and gated in an efficient cascade pipeline. 
	
	\item  An original tracking head LGCH is introduced. It further utilizes the HFC-expanded features by mapping the features to a higher-dimensional nonlinear feature space, whereby gating is performed through the Hadamard product. 
 
        \item  We perform comprehensive evaluations on three authoritative UAV tracking benchmarks demonstrating the state-of-the-art performance of CGTrack.
\end{itemize}

\begin{figure}[htbp] 
	\centering
	\includegraphics[width=0.98\linewidth]{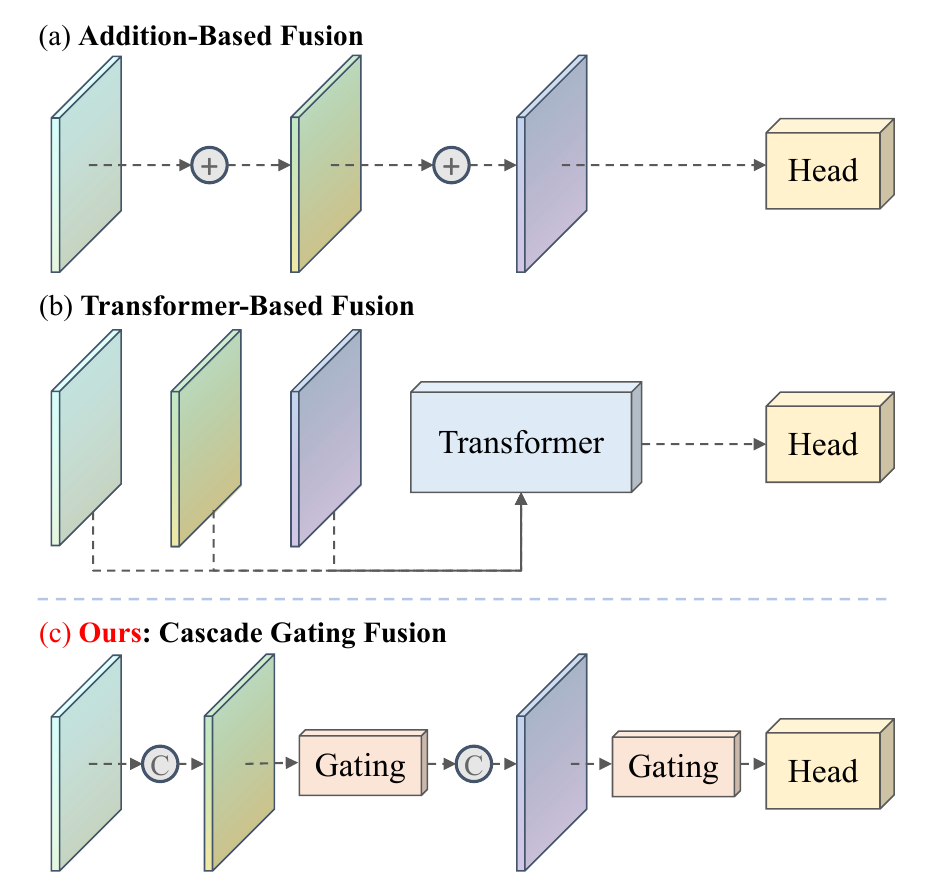}
	\setlength{\abovecaptionskip}{5pt}
	\caption
	{
            Comparison of the popular hierarchical feature fusion methods for UAV tracking. \textbf{(a)} Addition-Based Fusion: simply adds all the feature maps up. \textbf{(b)} Transformer-Based Fusion: employs Transformer layers or multi-head attention modules for feature fusion. \textbf{(c)} Cascade Gating Fusion: concatenates adjacent feature maps and performs gating subsequently in a cascade architecture.
	}
	\label{fig:fusion_vs}
	\vspace{-4mm}
\end{figure}

% \vspace{-2mm}
\section{Related Works}
% \vspace{-1mm}
\subsection{Visual Object Tracking for UAV.}
Despite the high efficiency of DCF-based trackers~\cite{kcf,danelljan2017eco,ATOM,DiMP,PrDiMP,KYS}, they have been largely supplanted by Siamese-based trackers~\cite{SiameseFC,SiameseRPN,SiamBAN,Stark,TransT,ostrack,SeqTrack} in UAV tracking due to their relatively low accuracy. More recently, some studies have attempted to incorporate Transformer~\cite{2017Attention} into Siamese-based UAV tracking pipeline~\cite{hift,SGDViT,ClimRT} to enhance the interaction between extracted template and search region features. However, these methods still adopt the two-stream framework, leading to insufficient information interaction during feature extraction. In this work, we adapt a lightweight hierarchical ViT into the one-stream UAV tracking pipeline, establishing an optimal equilibrium between computational demands and tracking accuracy

% \vspace{-1mm}

\subsection{Hierarchical UAV Trackers.}
In UAV scenarios, most existing trackers adopt lightweight image classification networks as backbones for better computational efficiency~\cite{hift,SiameseFC,siamapn++,SGDViT,ClimRT,tctrack,tctrack++}. However, the high-stride downsampling in these networks often leads to a loss of critical information. To mitigate this, hierarchical UAV trackers attempt to leverage multi-level feature maps generated at different stages of the lightweight backbones. For instance,  SiamAPN++~\cite{siamapn++} employs an attention mechanism to adaptively fuse multi-level features, while HiFT~\cite{hift} stacks Transformer layers to incorporate multi-scales feature maps. These methods, as shown in Fig.~\ref{fig:fusion_vs}(b), are burdened with heavy relation modeling. Beyond UAV-specific trackers, HiT~\cite{HiT} employs simple addition for hierarchical feature aggregation, also depicted in Fig.~\ref{fig:fusion_vs}(a). This approach, despite its simplicity, ignores the diverse variance of hierarchical feature maps, resulting in severe information loss. As shown in Fig.~\ref{fig:fusion_vs}(c), unlike the aforementioned methods, we propose a novel cascade gating structure that combines gating mechanism and feature reuse to expand network capacity with minor computational costs.   
% \vspace{-1mm}

\subsection{Gating Mechenism.}
Recent studies have demonstrated the utility of the gating mechanism across numerous computer vision tasks~\cite{SENet,deepfillv2,moganet,StarNet}. To illustrate, SENet~\cite{SENet} introduces an efficient channel-wise attention mechanism through lightweight gating. DeepFill v2~\cite{deepfillv2} incorporates the gating mechanism into the convolution to better distinguish between different pixels in an image. Moreover, StarNet~\cite{StarNet} further explains the reason why numerous efficient network designs adopt gating mechanisms~\cite{SENet,moganet}: the Hadamard product has the ability to map the features into higher and nonlinear dimensions, but operates in low-dimensional space. In this work, we integrate the coarse-gained gating and fine-grained gating into the tracking framework respectively to mine local discriminative details with efficiency.

\begin{figure*}[htbp]
	\centering
	\includegraphics[width=1.0\linewidth]{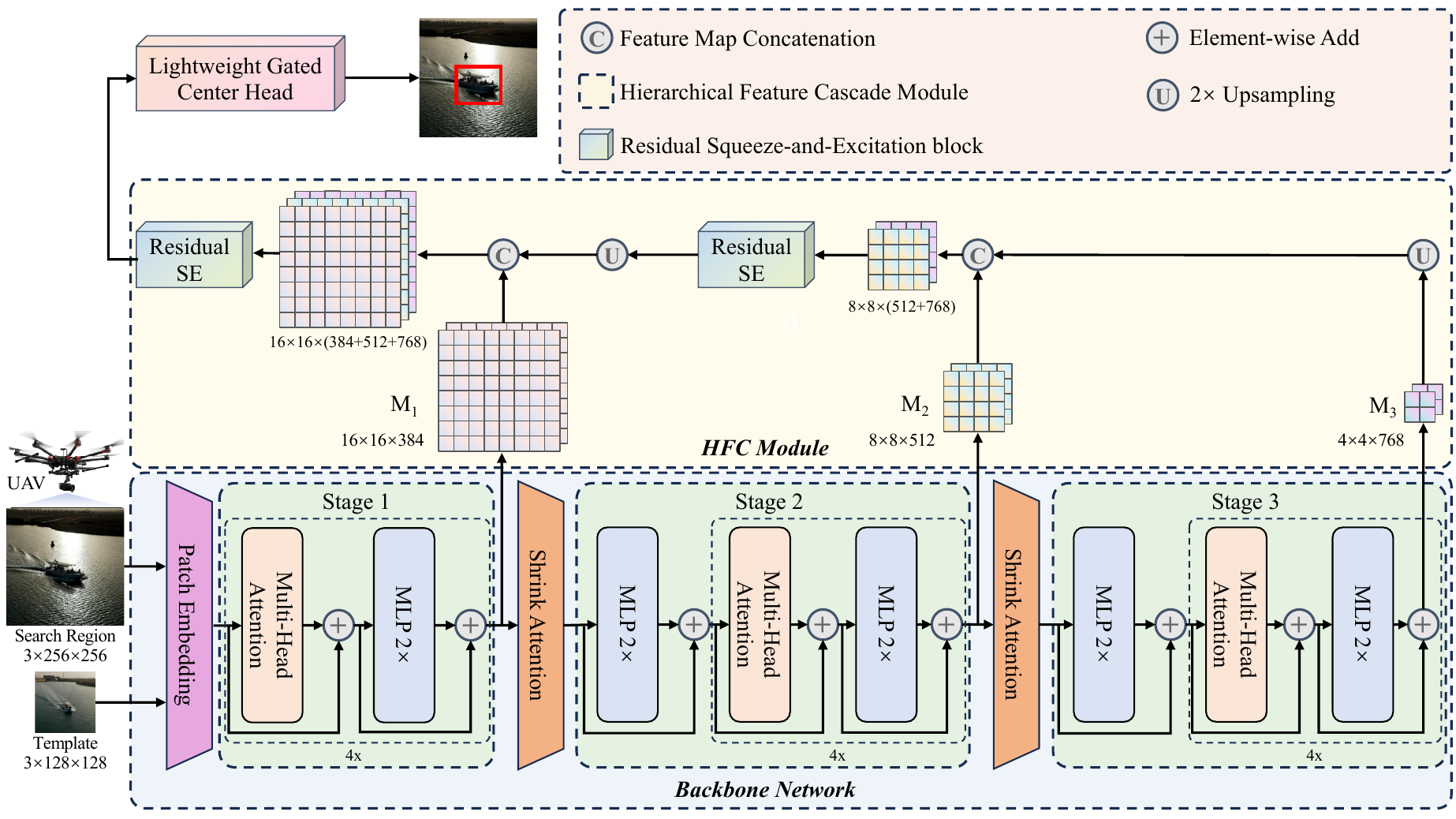}
	\caption{Overview of the proposed CGTrack, which comprises three main components: a lightweight hierarchical backbone, an HFC module, and a Lightweight Gated Center Head.
        }
	\label{fig:arch}
	\vspace{-4mm}
\end{figure*}

% \vspace{-0.5pt}
\section{Proposed Method}
This section systematically elaborates on our proposed CGTrack framework. We first establish a conceptual schematic of the architecture, followed by the detailed component introduction, including a lightweight ViT backbone, the proposed \textit{Hierarchical Feature Cascade} module, and \textit{Lightweight Gated Center Head}. In the final part of this section, we introduce the training objective.

\subsection{Overview}
As depicted in Fig.~\ref{fig:arch}, CGTrack is a one-stream tracking framework incorporating three core components: a lightweight ViT backbone, the proposed HFC module, and LGCH. Like most one-stream trackers~\cite{HiT,ostrack,ARTrack,SeqTrack}, CGTrack takes a pair of images as input and then jointly performs feature extraction and relational modeling across different network stages. The model progressively generates hierarchical feature maps with varying spatial resolutions from each ViT stage. Besides the backbone is a coarse-to-fine architecture consisting of our proposed HFC module and LGCH: The feature sequence is firstly fed into the HFC module for efficient feature augmentation and preliminary gating; Then, LGCH takes the expanded feature map as input and performs final purification to obtain tracking result.

\subsection{LeViT Backbone}
Inspired by HiT~\cite{HiT}, we adopt LeViT~\cite{graham2021levit} as the backbone of CGTrack and adapt it into the tracking framework. For clarification, we denote the input template image and search region image as $\mathbf{Z} \in {\mathbb{R}}^{3 \times {H_{z}} \times {W_{z}}}$ and $\mathbf{X} \in {\mathbb{R}}^{3 \times {H_{x}} \times {W_{x}}}$ respectively. They are first downsampled by a factor of 16 through patch embedding resulting in $\mathbf{Z_{p}} \in {\mathbb{R}}^{C \times {\frac{H_{z}}{16}} \times {\frac{W_{z}}{16}}}$ and $\mathbf{X_{p}} \in {\mathbb{R}}^{C \times {\frac{H_{x}}{16}} \times {\frac{W_{x}}{16}}}$. Then we flatten and concatenate $\mathbf{Z_{p}}$ and $\mathbf{X_{p}}$ in the spatial dimension and feed them into the following ViT stages. The transformer part of LeViT comprises three stages, and each stage consists of $Li$ blocks, \textit{i.e.}, $L1$$=$$4$, $L2$$=$$4$, $L3$$=$$4$. Each block has a Multi-Head Attention and an MLP in the residual form. LeViT leverages the Shrink Attention modules to downsample feature maps at a scale of 4 between stages, producing three feature maps with multiple resolutions. As is common in one-stream trackers, we extract the search region part of the output from each stage and re-interpret these tokens to a 2D spatial correlation map. Finally, we obtain a sequence including three correlation maps with distinct size, \textit{i.e.}, $\mathbf{M_{1}} \in {\mathbb{R}}^{{H_{s}} \times W_{s} \times C_{s}}$, $\mathbf{M_{2}} \in {\mathbb{R}}^{{H_{m}} \times W_{m} \times C_{m}}$, $\mathbf{M_{3}} \in  {\mathbb{R}}^{{H_{l}} \times W_{l} \times C_{l}}$, where ${C_{s}}=384$, ${C_{m}}=512$, ${C_{l}}=768$. In addition, a similar position encoding design, analogous to Dual-image Position Encoding in HiT is used in our CGTrack, to better adapt LeViT for the tracking task. Further details regarding the backbone network of CGTrack can be found in LeViT and HiT.

\Remark Attributing to the lightweight hierarchical ViT, we obtain a sequence of correlation maps preserving rich global context information in both search region and template with minor costs. This highly parallelized one-stream structure is able to handle multiple UAV scenarios with flexible variants.

\subsection{Hierarchical Feature Cascade Module}
Inspired by DenseNet~\cite{densenet}, we propose a Hierarchical Feature Cascade module that progressively integrates multi-stage backbone features via concatenation operations. By applying gating to the concatenated feature maps, the HFC module enhances the critical local discriminative details. Different from the original dense connection in DenseNet, the HFC module simplifies it by only keeping one cascade path between adjacent feature maps, which significantly reduces memory usage and achieves promising efficiency. As illustrated in Fig.~\ref{fig:arch}, for hierarchical 3D correlation maps denoted as $M_i,~i \in\{1, 2, 3\}$, we first upsample $M_1$ and concatenate it and $M_2$ together along the channel dimension, which can be written as 
\begin{equation}
\begin{split}
    {X} = {\rm{Concat}}({\mathbf{M_{2}}},{\rm{Upsample}}({\mathbf{M_{1}}}))
\end{split}
\label{eq-hfc}
\end{equation}
where $X$ is the intermediate result of the HFC module. In view of the diverse variance of the concatenated features, we adapt the original Squeeze-and-Excitation block~\cite{SENet} into the residual form and propose Residual SE, which efficiently applies channel re-scaling and preliminary gating to the concatenated feature map through Hadamard product. The entire process of Residual SE can be formulated as 
\begin{gather}
        % Squeeze operation
        z_c = \frac{1}{H \times W} \sum_{i=1}^{H} \sum_{j=1}^{W} x_{c, i, j} \\
        % Excitation operation
        s_c = \sigma(W_2 \cdot \text{ReLU}(W_1 \cdot z_c)), \\
        % Scale operation
        \hat{x}_{c, i, j} = x_{c, i, j} \cdot s_c, \\
        % Comprehensive formula
        \hat{X} = X + X \odot S 
\end{gather}
where $x_{c, i, j}$ represents the input feature at channel $c$ and spatial location $(i, j)$. $z_c$ is the channel descriptor obtained via global average pooling. $W_1$ and $W_2$ are the weights of the fully connected layers used in the excitation step. $\sigma$ is the sigmoid activation function. $X$ and $\hat{X}$ are the overall input and output feature maps, respectively. $\odot$ denotes Hadamard product. Then, we apply the same operation to the output of the previous step to obtain the final feature. The entire process can be mathematically formulated as:
\begin{align} 
    % step 1
    {O} &= {\rm{ResidualSE}}({\rm{Concat}}({\mathbf{M_{2}}},{\rm{Upsample}}({\mathbf{M_{1}}}))) \\
    % step 2
    {Y} &= {\rm{ResidualSE}}({\rm{Concat}}({\mathbf{M_{3}}},{\rm{Upsample}}({\mathbf{O}})))
\end{align}

\Remark
Compared to additive operations, feature concatenations preserve all information and improve gradient flow, thereby accelerating network convergence. Each concatenation operation yields a higher-dimensional feature map explicitly. By applying the Residual SE to the concatenated feature map, we enhance local discriminative details through gating. This amplification is particularly crucial in challenging UAV scenarios, such as tracking small objects or handling sudden appearance changes.

\begin{figure}[t] 
	\centering
	\includegraphics[width=1\linewidth]{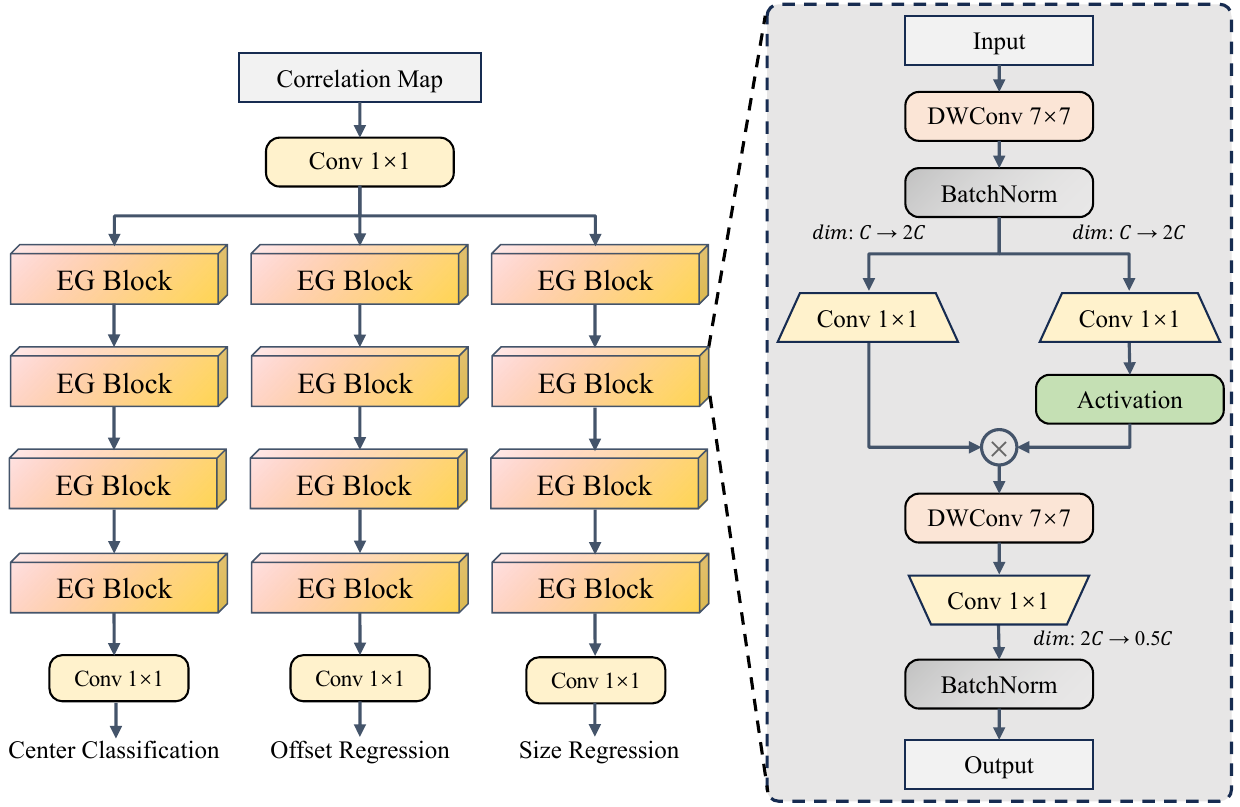}
        \caption{Detailed architectures of LGCH. The left part illustrates the overall workflow of LGCH. The right one shows the structure of the EG block.}
	\label{fig:LGCH}
	\vspace{-6mm}
\end{figure}

\subsection{Lightweight Gated Center Head}
From the HFC module, we obtain features enriched with contextual information. Building on this, we propose the Efficient Gating block to extract fine-grained discriminative details from the expanded features. As illustrated in Fig.~\ref{fig:LGCH}, the concatenated feature map is first downsampled by a 1 × 1 convolution into a channel size of 256 for memory efficiency. The feature map is then fed into three branches, each containing four EG blocks followed by a 1 × 1 convolution. Consistent with ~\cite{ostrack}, the output of three branches is a classification score map, a local offset map, and a bounding box size map, respectively. The detailed structure of the EG block is depicted on the right side of Fig.~\ref{fig:LGCH}. The core design of the EG block involves two 1 × 1 convolutions that map the input into higher-dimensional, non-linear feature space. Among these, one branch incorporates an activation function, forming the $gate$ branch while the other serves as the $context$ branch. Subsequently, the Hadamard product is performed between $gate$ and $context$. The entire process of the EG block can be written as:
\begin{align}
    O &= \mathrm{BN}(\mathrm{DW}_{7\times 7}(X)) \\
    X_1 &= \mathrm{Conv}_{1\times 1}(O) \\
    X_2 &= \mathrm{Conv}_{1\times 1}(O) \\
    P &= \mathrm{ReLU6}(X_1) \odot X_2 \\
    Y   &= \mathrm{BN}(\mathrm{Conv}_{1\times 1}(\mathrm{DW}_{7\times 7}(P)))
\end{align}    
where $X_1$ and $X_2$ are the $gate$ and $context$ branch, respectively.

\Remark 
Unlike commonly used CBR blocks, the proposed EG blocks exhibit enhanced capability in extracting fine-grained features. Similar to the HFC module, EG blocks first explicitly map features to higher dimensions before performing gating, effectively decoupling discriminative, target-oriented information in the correlation map. Furthermore, as described in ~\cite{StarNet}, the Hadamard product implicitly transforms previously expanded features into exceptionally high and nonlinear dimensions while maintaining operations in a low-dimensional space. Notably, our proposed EG block has even fewer FLOPs and parameters than a CBR block, highlighting its potential for applications on edge devices.

\begin{figure*}[!t]	
    \centering
	\includegraphics[width=0.3\linewidth]{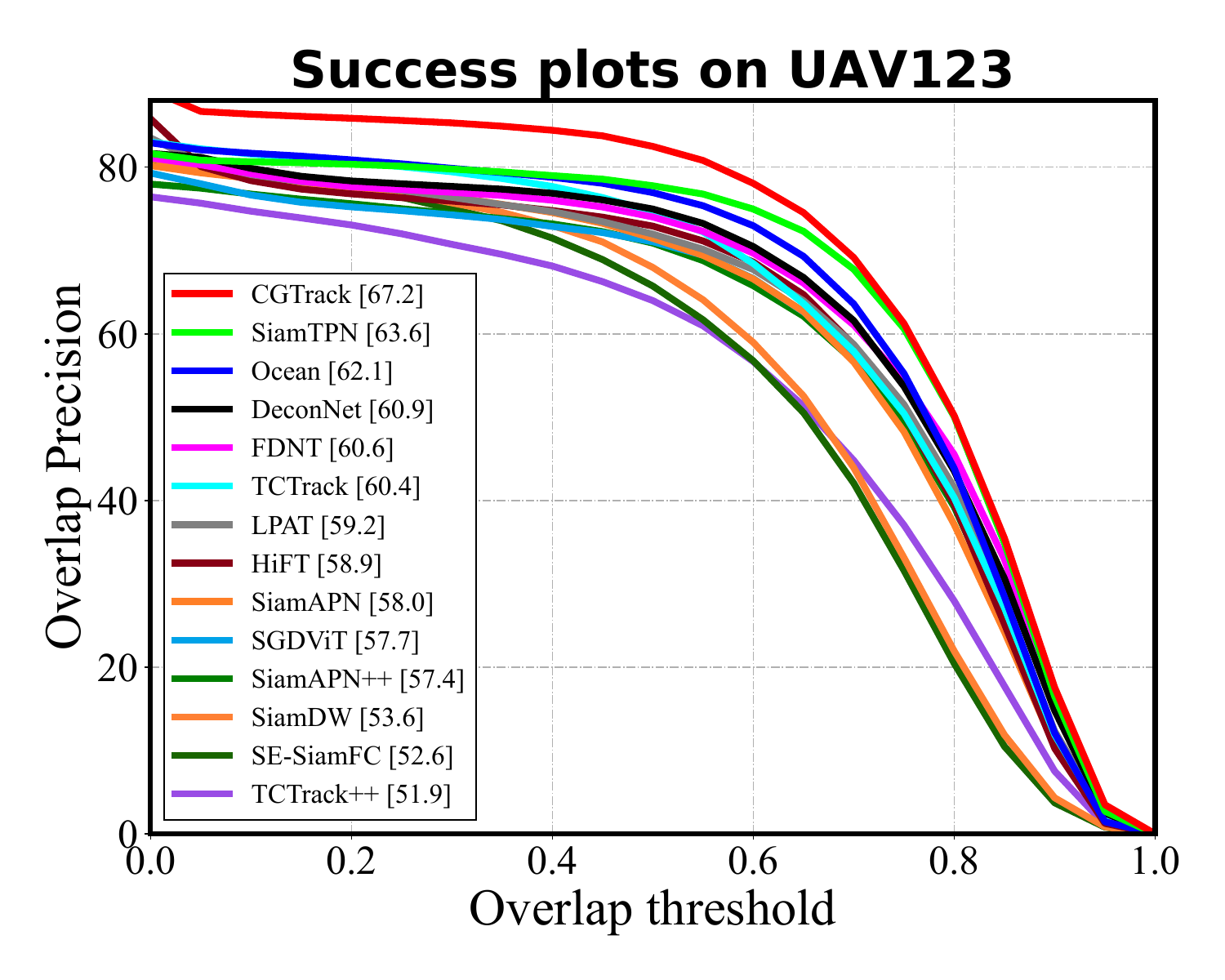}
	\includegraphics[width=0.3\linewidth]{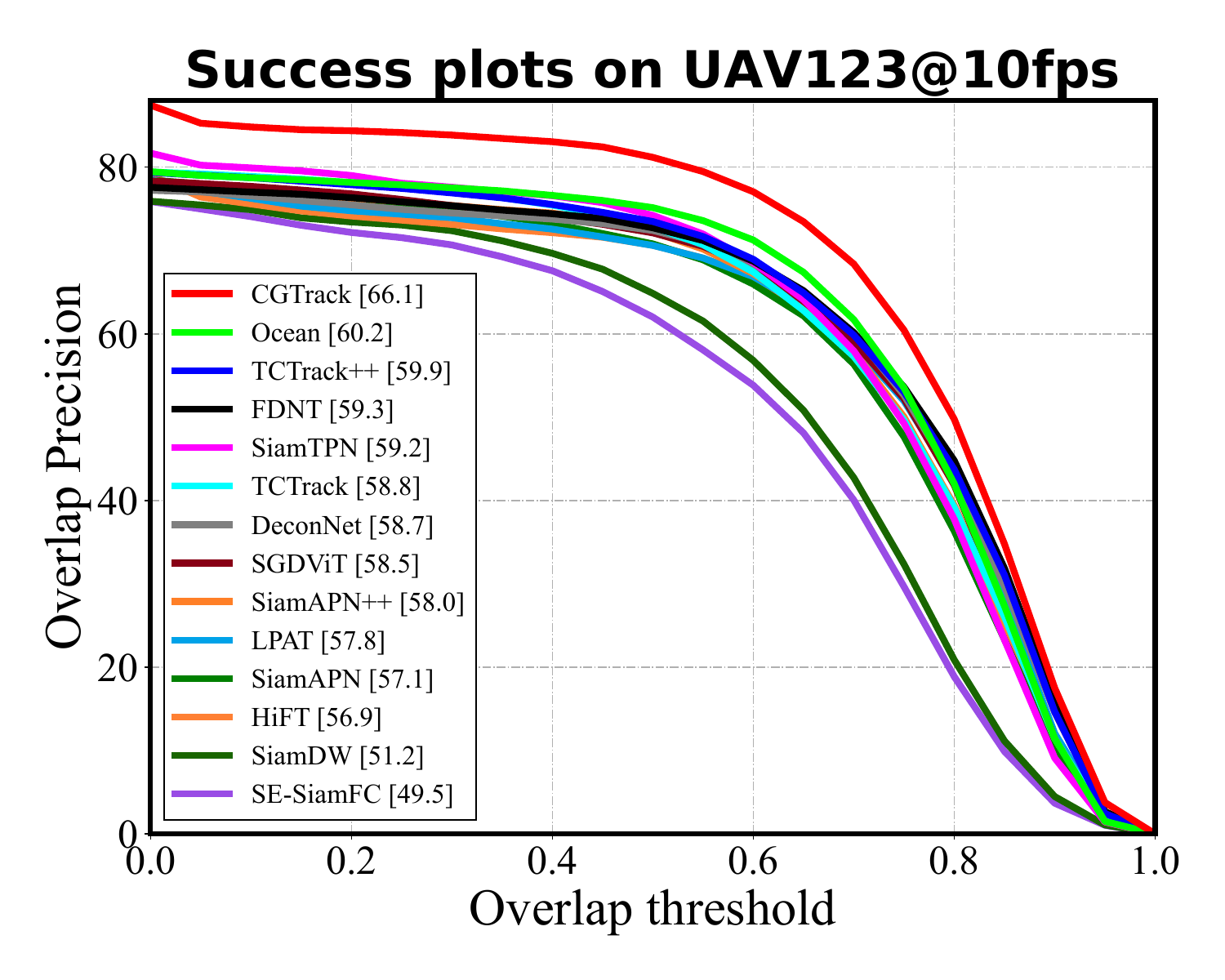}
	\includegraphics[width=0.3\linewidth]{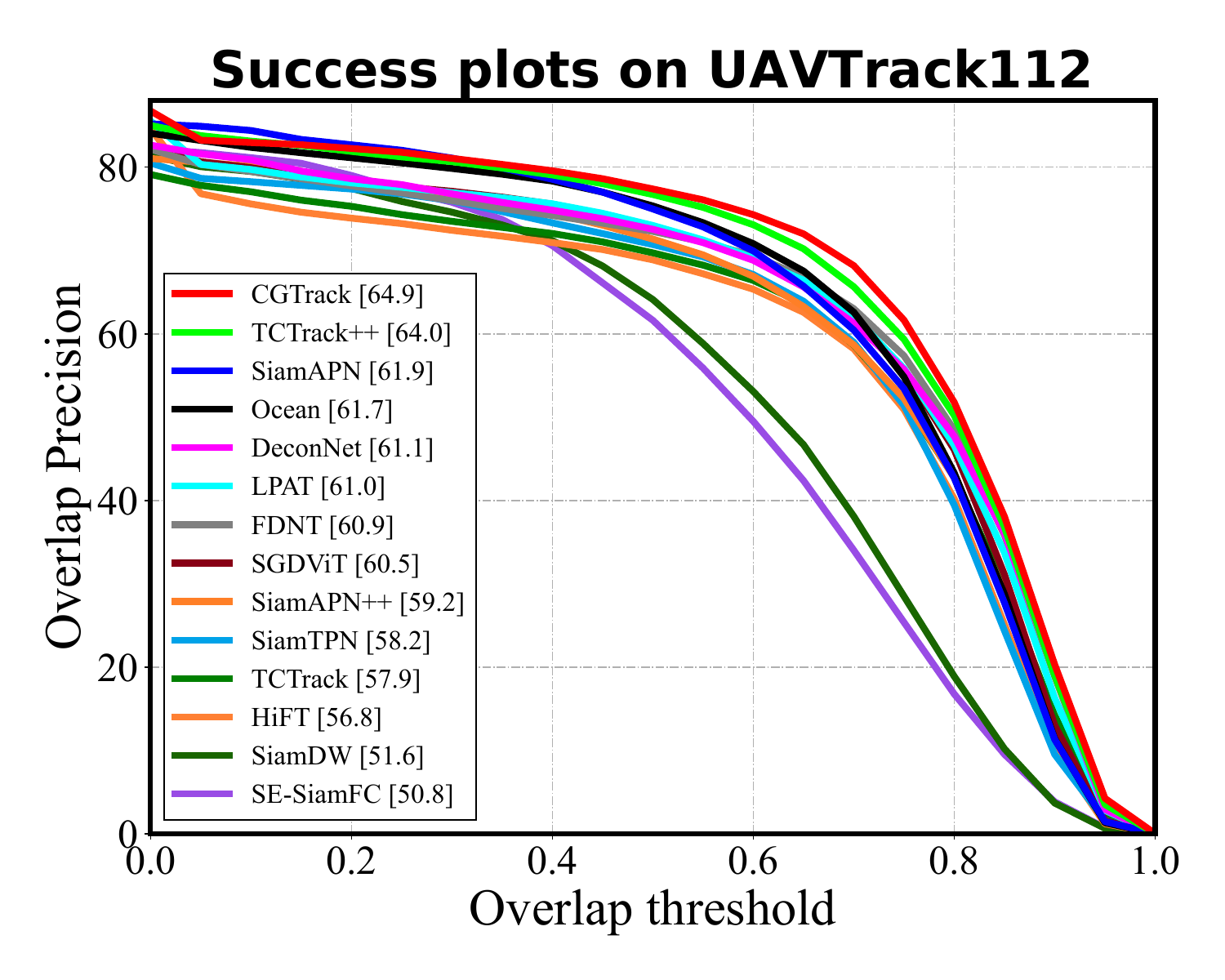}
    \label{fig:all-1}
    \vspace{-0.6cm}
\end{figure*}
\begin{figure*}[!t]	
    \centering
	\includegraphics[width=0.3\linewidth]{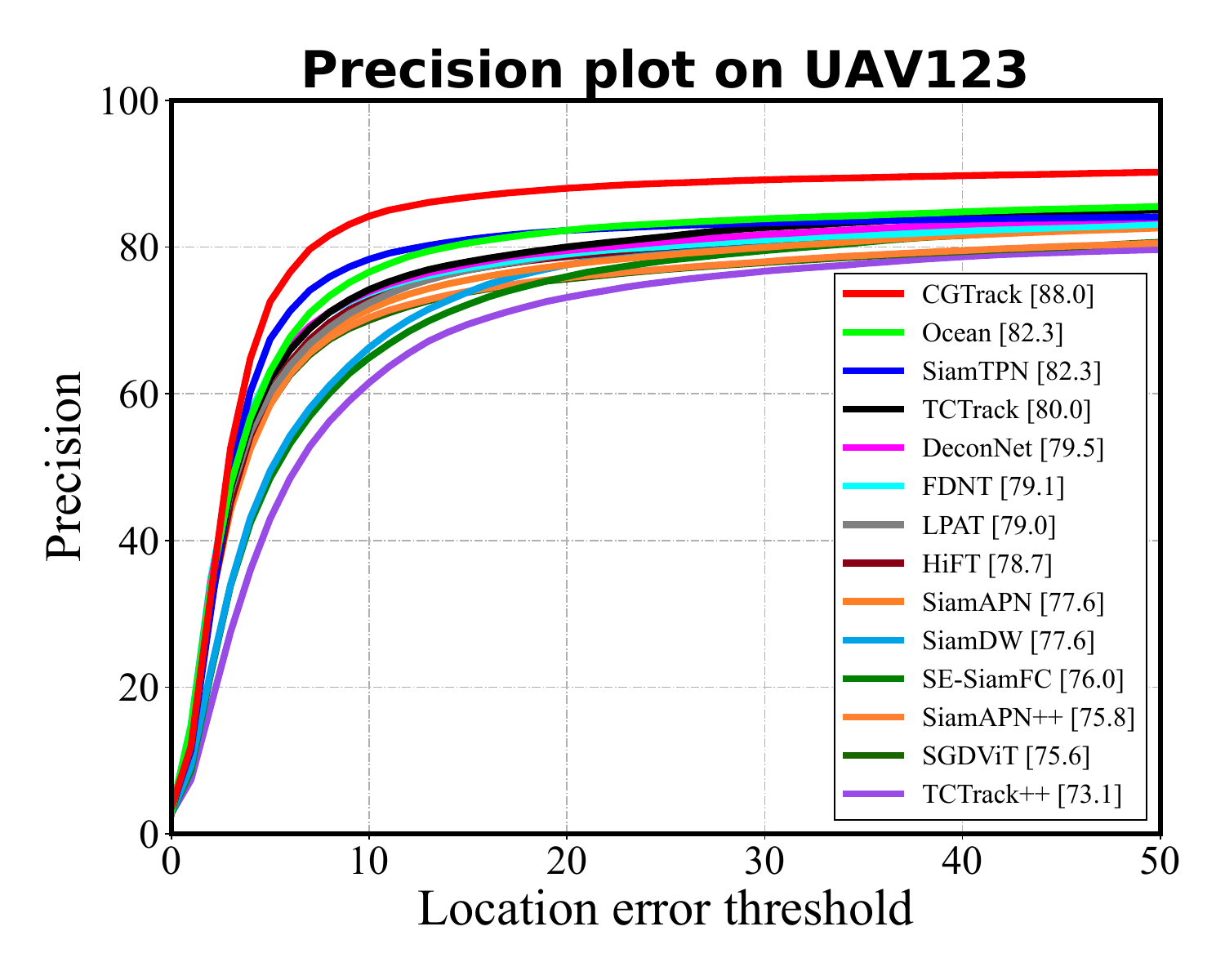}
	\includegraphics[width=0.3\linewidth]{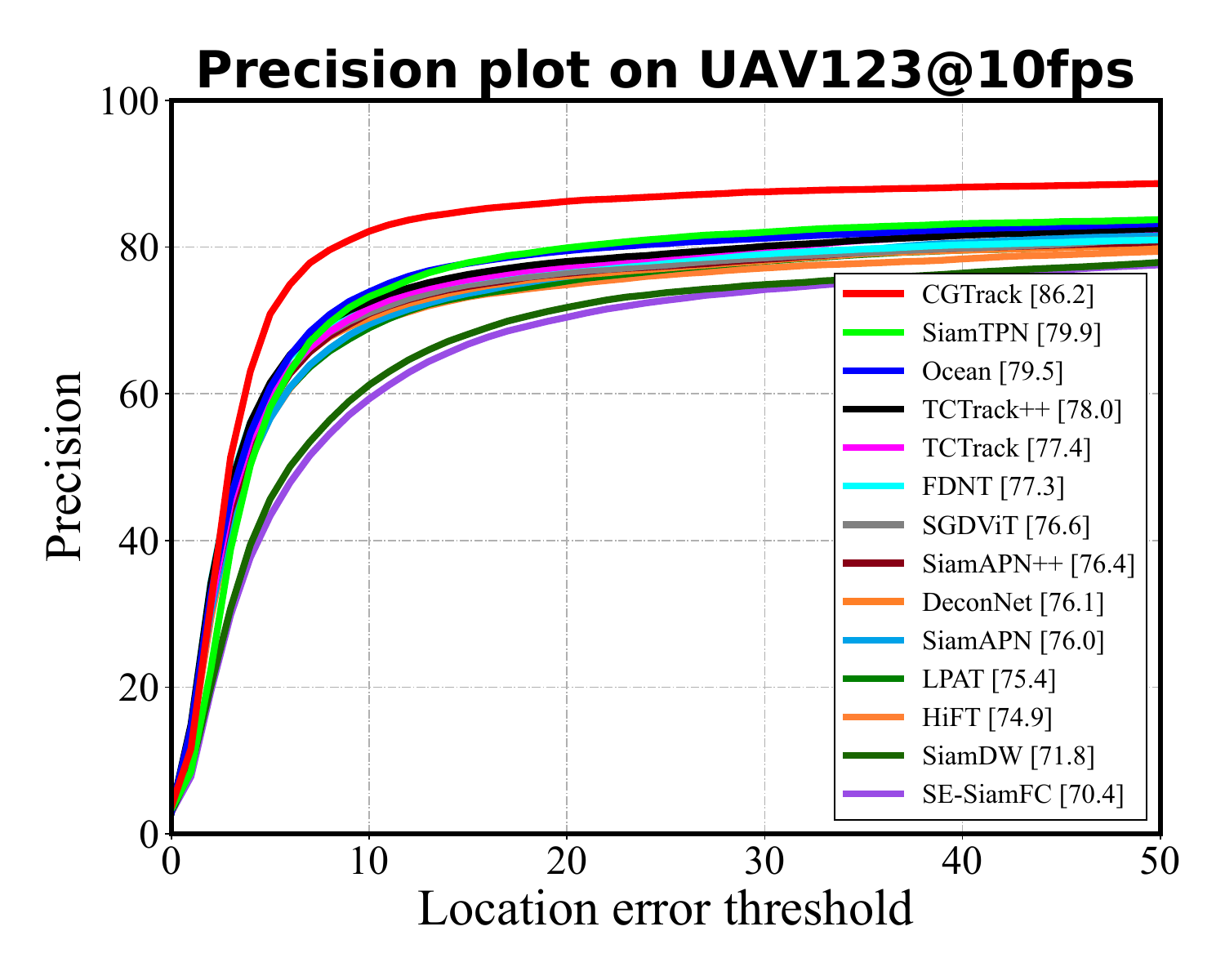}
	\includegraphics[width=0.3\linewidth]{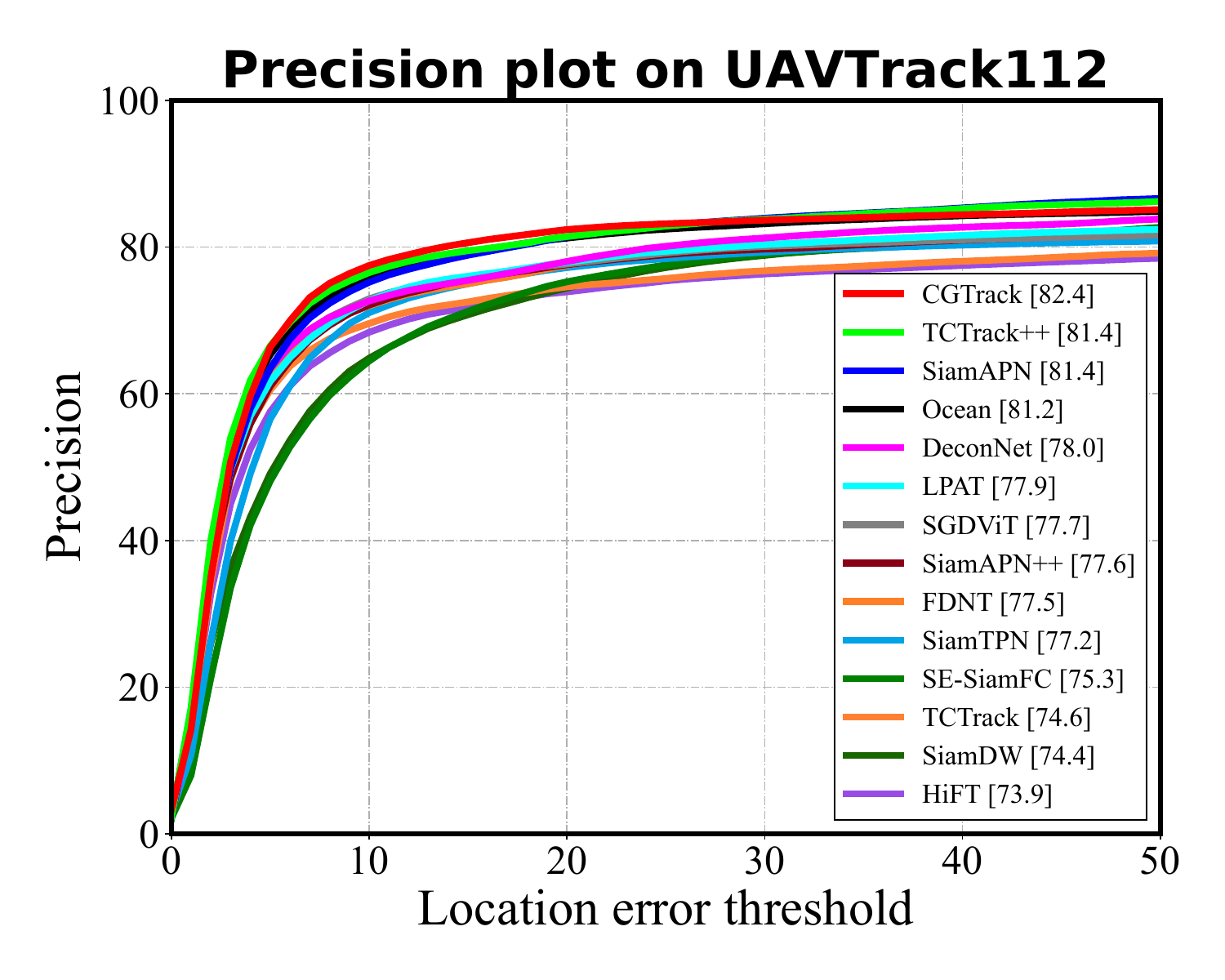}
    \caption
	{
        Overall performance of CGTrack and prevailing SOTA trackers on UAV123 \cite{uav123} (the first column), UAV123@10fp \cite{uav123} (the second column), and UAVTrack112 \cite{uav112} (the third column) benchmarks. CGTrack achieves SOTA performance across all benchmarks.
	}
    \label{fig:all}
    \vspace{-0.4cm}
\end{figure*}

\subsection{Training objective}
\label{sec:training_and_inference}
In the training phase, we employ weighted focal loss~\cite{focal} in the classification task, while utilizing a combination of $\ell_1$ loss and generalized GIoU loss~\cite{GIoU} for the localization task.
The overall loss function is
\begin{equation}
	\label{equ-loss-loc}
	\begin{aligned}
		\mathcal{L}=\mathcal{L}_{\operatorname{focal}} + \lambda_{G}\mathcal{L}_{\operatorname{GIoU}}+\lambda_{l}\mathcal{L}_l,
	\end{aligned}
\end{equation}
where $\lambda_{G}=2$ and $\lambda_{l}=5$ are the regularization parameters following ~\cite{ostrack}.

\section{Experiments}
\subsection{Implementation Details}
The proposed CGTrack is implemented in Python 3.8 with PyTorch 1.10.0, trained on four NVIDIA RTX 3090 GPUs. We employ the train-splits of GOT-10k~\cite{GOT10K} (excluding 1k sequences as convention), LaSOT~\cite{LaSOT}, COCO2017~\cite{COCO}, and TrackingNet~\cite{trackingnet} for training. Common data augmentations including horizontal flipping and brightness jittering are applied during training. The network processes 128$\times$128 template and 256$\times$256 search images in each training batch of 128 samples.  We adopt AdamW~\cite{AdamW}, with the weight decay of 1e-4 as the optimizer. The initial learning rate of CGTrack is set to 4e-5 which decays by 10\% during the final 20\% training epochs.

% In this work, the proposed CGTrack is implemented in Python 3.8 using PyTorch 1.10.0. Models are trained on 4 NVIDIA RTX 3090 GPUs. We adopt the train-splits of GOT-10k~\cite{GOT10K}(without 1k sequences as convention), LaSOT~\cite{LaSOT}, COCO2017~\cite{COCO}, TrackingNet~\cite{trackingnet} for training. Common data augmentations including horizontal flip and brightness jittering are employed in training. The total batch size is set at 128. The search and template images are resized to 256 $\times$ 256 and 128 $\times$ 128, respectively. We adopt AdamW~\cite{AdamW}, with the weight decay of 1e-4 as the optimizer. The initial learning rate of CGTrack is set to 4e-5. We reduce the learning rate by 10\% in the last 20\% epochs.

\subsection{Overall Performance on UAV Tracking Benchmarks}
In this subsection, our CGTrack is comprehensively compared with 13 SOTA trackers including TCTrack~\cite{tctrack}, SGDViT~\cite{SGDViT}, FDNT~\cite{fdnt}, HiFT~\cite{hift}, SiamAPN~\cite{uav112}, LPAT~\cite{lpat}, DeconNet~\cite{deconnet}, SiamTPN~\cite{siamtpn}, SiamDW~\cite{Deeper-wider-SiamRPN}, TCTrack++~\cite{tctrack++}, SiamAPN++~\cite{siamapn++}, SE-SiamFC~\cite{sesiamfc}, Ocean~\cite{Ocean} on three public authoritative aerial tracking benchmarks.

\textit{UAV123.} UAV123~\cite{uav123} is a comprehensive aerial video benchmark dataset containing 123 HD sequences with over 112,000 frames captured from low-altitude aerial platforms. This benchmark includes various challenging scenarios such as rapid target motion and scale variation, providing a comprehensive platform to thoroughly evaluate CGTrack's performance in aerial tracking. As depicted in Fig.~\ref{fig:all}, CGTrack demonstrates SOTA tracking performance with 88.0\% in Precision and 67.2\% in Success score.
% CGTrack significantly outperforms other SOTA trackers in both Precision (88.0\%) and Success score (67.2\%).

\textit{UAV123@10fps.} UAV123@10fps~\cite{uav123} is derived by downsampling the original 30fps version, which leads to more pronounced motion between consecutive frames. This increased motion poses a challenge to trackers in more effectively leveraging inter-frame continuity information for robust aerial tracking. As shown in Fig.~\ref{fig:all}, CGTrack consistently achieves SOTA performance, with the highest Precision (83.8\%) and Success score (66.1\%).

\textit{UAVTrack112.} UAVTrack112~\cite{uav112} is a challenging aerial benchmark that collects 112 real-world sequences including low illumination scenarios in the dark time. As illustrated in Fig.~\ref{fig:all}, CGTrack sets a new SOTA Success score of 64.9\% and Precision score of 80.6\%.

\begin{figure}[!t] 
	\centering
	\includegraphics[width=0.98\linewidth]{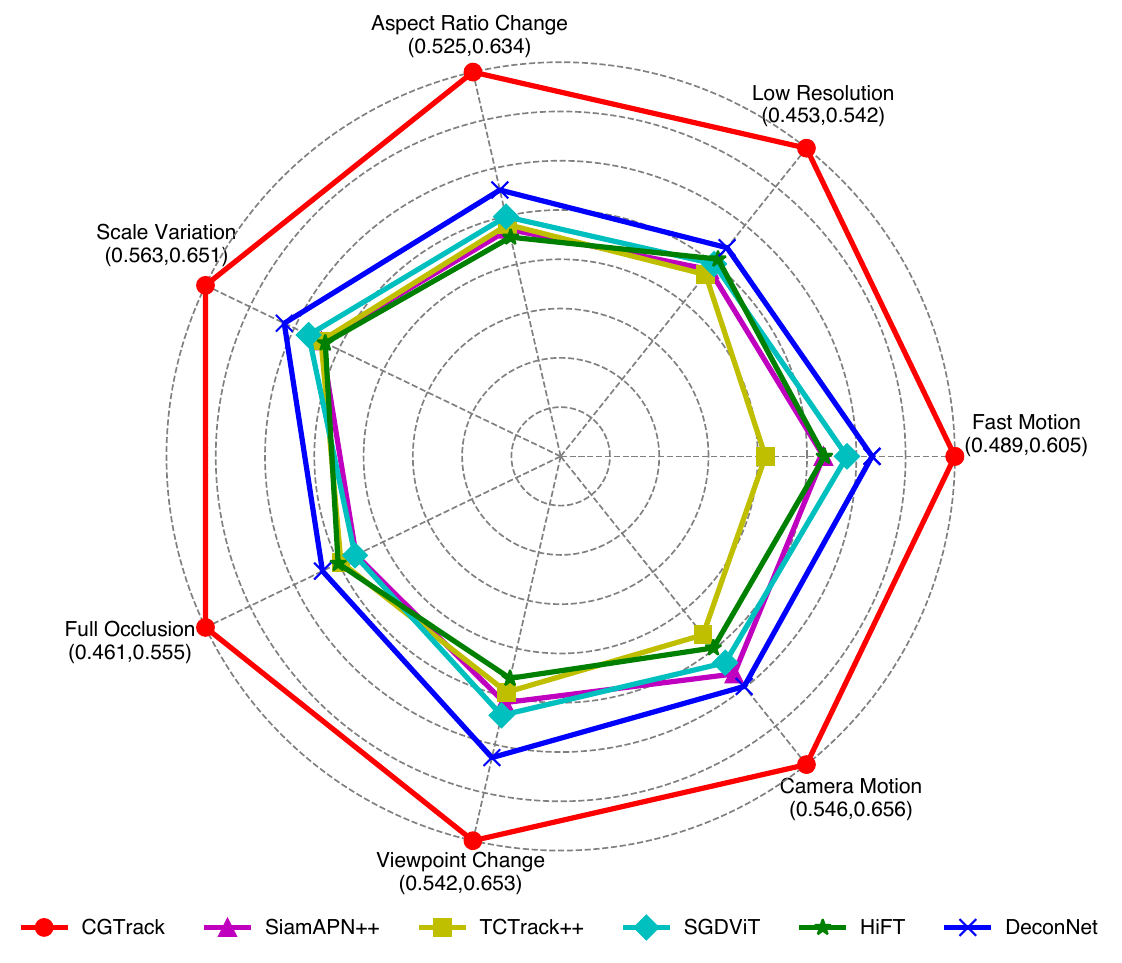}
	\setlength{\abovecaptionskip}{-1pt}
	\caption
	{
            Success scores of different attributes among top 6 SOTA UAV trackers. CGTrack significantly outperforms other trackers in typical UAV attributes.
	}
	\label{fig:attr}
	\vspace{-2mm}
\end{figure}

\subsection{Attribute-Based Comparison}
To further evaluate the robustness of CGTrack against diverse UAV-specific challenges, we conduct exhaustive attribute-based comparisons with other 5 SOTA UAV trackers. As depicted in Fig.~\ref{fig:attr}, our CGTrack achieves SOTA performance in all attributes. The promising results demonstrate that CGTrack is capable of aggregating the mined local discriminative details and global context to mitigate various challenges in UAV scenarios. 

\begin{table}[!htbp]
    \centering
    \resizebox{\linewidth}{!}{
        \begin{tabular}{c|l|ccc}
            \toprule
            \# & \multicolumn{1}{c|}{Method}         & Precision   & Params(M) & MACs(G)\\
            \midrule[0.5pt]
            1  & Addition-based Fusion                     & 82.90 & 40.931      &  4.315  \\

            2  & Concatenation-based Fusion(w/o Residual SE)  & 84.12 & 40.668     &  4.323   \\

            3  & \myAbl Concatenation-based Fusion + Residual SE & \myAbl \textbf{86.24} &\myAbl {41.219}   & \myAbl {4.324}   \\
            \bottomrule
        \end{tabular}}
        \setlength{\abovecaptionskip}{4pt}
        \caption{Ablation study of different fusion manners of CGTrack. \colorbox{gray!18}{Gray} color is employed to denote our final configuration.}
        \label{tab:ab-1}
        \vspace{-3mm}
\end{table} 
\begin{table}[!htbp]
    \centering
    \resizebox{\linewidth}{!}{
        \begin{tabular}{c|c|cccc}
            \toprule
            \# & \multicolumn{1}{c|}{Method}                & Params(M) & MACs(G) & FPS   &  Precision   \\
            \midrule[0.5pt]
            1  & CGTrack-T                  & 9.987  & 1.165   &  61.4  & 80.08 \\

            2  & CGTrack-S                  & 11.421  & 1.300  &  55.6  & 82.00  \\

            3  & CGTrack-B                  & 41.219  & 4.324  &  42.1  & 86.24 \\
            \bottomrule
        \end{tabular}}
        \setlength{\abovecaptionskip}{4.5pt}
        \caption{Details and variants of our CGTrack model.}
        \label{tab:ab-2}
        \vspace{-2mm}
\end{table} 
\begin{table}[!htbp]
    \centering
    \resizebox{\linewidth}{!}{
        \begin{tabular}{c|c|ccc}
            \toprule
            \# & \multicolumn{1}{c|}{Method}                & Head Params(M)    & MACs(G)     & AUC   \\
            \midrule[0.5pt]
            1  & Plain CBR block              & 2.935  & 0.751      & 65.51    \\

            2  & EG block-1x                  & 1.425  & 0.364      & 63.64    \\

            \myAbl  3  & \myAbl EG block-2x    & \myAbl \textbf{2.665}  & \myAbl \textbf{0.680}    & \myAbl \textbf{66.14}   \\

            4  & EG block-3x                  & 3.904  & 0.996      & 65.31    \\

            4  & EG block-4x                  & 5.143  & 1.313      & 63.40    \\
            \bottomrule
        \end{tabular}}
        \setlength{\abovecaptionskip}{5pt}
        \caption{Ablation study on different components and configurations of the center head. \colorbox{gray!18}{Gray} denotes our final configuration.}
        \label{tab:ab-3}
        \vspace{-7mm}
\end{table} 

\subsection{Ablation Study and Visualization}
In this subsection, we present the ablation studies on UAV123@10fps.

\textit{Hierarchical Feature Fusion Analysis.}
To verify the superiority of our fusion manner, we compare different hierarchical feature fusion manners. As shown in Tab.~\ref{tab:ab-1}, the original concatenation-based fusion (without gating) gains a 1.22\% improvement in Precision score. Furthermore, as enumerated in Tab.~\ref{tab:ab-1}, Row 3, the cascade gating framework exhibits a 3.34\% increase in Precision score compared to Row 2. The aforementioned results demonstrate the efficacy of the designed HFC module in UAV tracking.

\textit{Variants Analysis.}
In Tab.~\ref{tab:ab-2}, we present multiple variants of CGTrack with different backbone networks. Specifically, we adopt LeViT-384~\cite{graham2021levit}, LeViT-128, and LeViT-128S for CGTrack-B, CGTrack-S, and CGTrack-T, respectively. Among these variants, CGTrack-T exhibits superior speed at 61.4 fps on an NVIDIA RTX 3090 GPU, while CGTrack-B focuses on robustness, achieving an 86.24\% Precision score on UAV123@10fps. Notably, the CGTrack-S achieves a more favorable balance between computational complexity and tracking performance. The designed variants show strong generalization across different application scenarios.

\textit{LGCH Analysis.}
This part analyzes the effectiveness of LGCH and compares different feature upsampling ratios in the EG block. As in Tab.~\ref{tab:ab-3}, LGCH outperforms the CBR-based center head with even fewer parameters and FLOPs. When setting the upsampling ratio to 2, the highest Success score is achieved. The results indicate that a larger upsampling ratio can cause overfitting, while a smaller upsampling ratio may result in insufficient modeling capacity.

\begin{figure}[t] 
	\centering
	\includegraphics[width=0.98\linewidth]{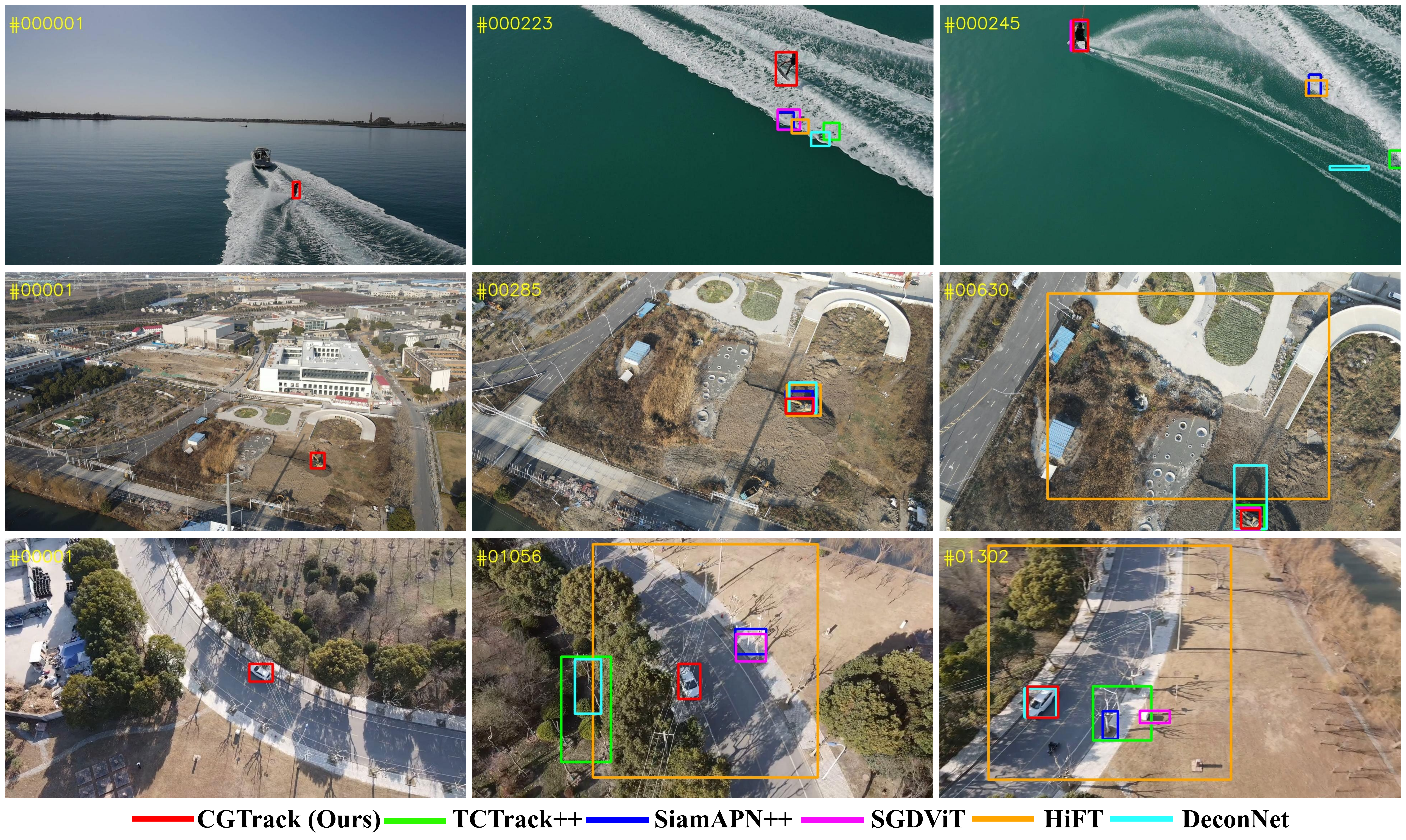}
	\setlength{\abovecaptionskip}{4pt}
	\caption
	{
            Qualitative comparison of CGTrack with other trackers on three representative sequences (\textit{wakeboard2} from UAV123@10fps, and \textit{excavator}, \textit{car4} from UAVTrack112). CGTrack achieves robust performance under severe UAV-specific challenges.
	}
	\label{fig:vis}
	\vspace{-7mm}
\end{figure}

\textit{Qualitative Results.} 
To intuitively demonstrate the tracking performance in real-world scenarios, we visualize the tracking results in Fig.~\ref{fig:vis}. The qualitative results across multiple real-world challenging scenes demonstrate that our CGTrack achieves superior robustness and accuracy, outperforming all the other UAV trackers. For further visualization of our method and comparison to other trackers, please kindly refer to the accompanying video.

\section{Conclusion}
In this work, we introduce a novel family of lightweight one-stream UAV trackers, dubbed CGTrack. CGTrack integrates global contextual information with mined local discriminative details to bridge the gap between lightweight ViTs and robust UAV tracking. By leveraging the art of feature reuse and gating mechanism, CGTrack significantly expands network capacity to tackle challenges in UAV scenarios without additional computational overhead. Extensive experiments demonstrate the superior real-world practicability and state-of-the-art performance of CGTrack. Finally, we hope this work could inspire and facilitate future research in robust UAV tracking. 

\section*{ACKNOWLEDGMENT}
Libo Zhang is supported by National Natural Science Foundation of China (No. 62476266). Heng Fan is not supported by any fund for this work.


%%%%%%%%%%%%%%%%%%%%%%%%%%%%%%%%%%%%%%%%%%%%%%%%%%%%%%%%%%%%%%%%%%%%%%%%%%%%%%%%

%%%%%%%%%%%%%%%%%%%%%%%%%%%%%%%%%%%%%%%%%%%%%%%%%%%%%%%%%%%%%%%%%%%%%%%%%%%%%%%%

%%%%%%%%%%%%%%%%%%%%%%%%%%%%%%%%%%%%%%%%%%%%%%%%%%%%%%%%%%%%%%%%%%%%%%%%%%%%%%%%
% \iffalse


% \section*{ACKNOWLEDGMENT}

% The preferred spelling of the word ÒacknowledgmentÓ in America is without an ÒeÓ after the ÒgÓ. Avoid the stilted expression, ÒOne of us (R. B. G.) thanks . . .Ó  Instead, try ÒR. B. G. thanksÓ. Put sponsor acknowledgments in the unnumbered footnote on the first page.
% \fi

%%%%%%%%%%%%%%%%%%%%%%%%%%%%%%%%%%%%%%%%%%%%%%%%%%%%%%%%%%%%%%%%%%%%%%%%%%%%%%%%
\bibliographystyle{IEEEtran}

\bibliography{main}
%\printbibliography

% %\begin{thebibliography}{100}
% %\bibliographystyle{IEEEtran}
% %\end{thebibliography}

\end{document}